%% file: 0_main.tex
\documentclass[letterpaper, 10 pt, conference]{ieeeconf}

\IEEEoverridecommandlockouts

\usepackage{etextools}
\usepackage{amssymb}
\usepackage{float}
\usepackage{makeidx}
\usepackage{amsmath}
\usepackage{amsfonts}
\usepackage{bbm}
\usepackage{epsfig}
\usepackage{epsf}
\usepackage{dsfont}
\usepackage{psfrag}
\usepackage{verbatim}
\usepackage{multirow}
\usepackage{tabularx}
\usepackage{booktabs}
\usepackage[tight,footnotesize]{subfigure}
\usepackage{array}
\usepackage{soul}
\usepackage{footnote}
\usepackage{cite}
\usepackage{dblfloatfix}

\usepackage{color, colortbl}
\usepackage[colorlinks,bookmarksnumbered,citecolor=orange,urlcolor=orange]{hyperref}
\usepackage{graphicx}
\graphicspath{{Figures}}
\DeclareGraphicsExtensions{.pdf,.png}
\usepackage{bigstrut}
\usepackage[english]{babel}

\usepackage{csquotes}

\usepackage[T1]{fontenc}
\usepackage{algorithm, setspace}
\usepackage{algpseudocode}
\usepackage{url}
\usepackage{multirow}
\usepackage{float}
\usepackage{xcolor}
\usepackage{hyperref}
 \hypersetup{
     colorlinks=true,
     linkcolor=orange,
     filecolor=orange,
     citecolor=orange,      
     urlcolor=orange,
     }

\let\labelindent\relax
\usepackage{enumitem}

\usepackage{xcolor}

\newcommand{\p}[1]{\smallskip \noindent \textbf{{#1}.}}
\newcommand{\eq}[1]{Equation~(\ref{eq:#1})}
\newcommand{\fig}[1]{Figure~\ref{fig:#1}}

\usepackage{graphicx}
\usepackage{cuted}
\usepackage{capt-of} 
\begin{document}

\title{CASTER: Formalizing Zero-Shot Manipulation \\ from Synthetic Videos as Constrained Optimization
}

\author{Yinlong Dai$^1$, Homanga Bharadhwaj$^2$, and Dylan P. Losey$^1$
\thanks{This work was supported in part by NSF Grant $\#2337884$. \newline $^1$\href{https://collab.me.vt.edu}{Collab}, Dept. of Mechanical Engineering, Virginia Tech, Blacksburg, VA. \texttt{\{yinlongdai, losey\}@vt.edu} \newline $^2$\href{http://b3.cs.jhu.edu/}{B3}, Dept. of Computer Science, Johns Hopkins University, Baltimore, MD.}
}

\maketitle

\begin{abstract}

Prior research on learning from humans often requires real human demonstrations.
But advances in video generation models offer a new paradigm: we can now prompt these generators to make \textit{synthetic videos} of the task.
On the plus side, these synthetic videos take place in the robot's current environment, and do not require actual human teaching.
However, while the generated videos appear to show plausible motions, they lack both physical consistency and robot grounding, and thus the robot cannot directly roll out the trajectory shown by the video.
In this paper we seek to formalize the process of going from synthetic video generation to real-world robot trajectories for zero-shot manipulation tasks.
We recognize that this problem is an instance of \textit{constrained optimization}.
The object trajectories extracted from the synthetic video provide an initial guess at the correct motion, and the robot must refine these trajectories (for instance, within a digital twin) while enforcing physical feasibility and semantic consistency.
Our optimization-based formalism provides insights into the video generation properties necessary for success.
Specifically, we show that the video generation must probabilistically seed the search within a semantic basin of attraction where the trajectory can be locally modified without losing its task alignment.
Our analysis further reveals that state-of-the-art approaches can be viewed as instances of our unified formalism with key simplifications.
In our experiments we remove these simplifications to more closely match the underlying optimization, leading to more robust zero-shot trajectories from synthetic videos and improving average task success by 34.1\% compared to the best-performing baseline.
See videos at our project website: \url{https://collab.me.vt.edu/CASTER}
\end{abstract}

\begin{figure*}[t]
    \centering
    \includegraphics[width=\textwidth]{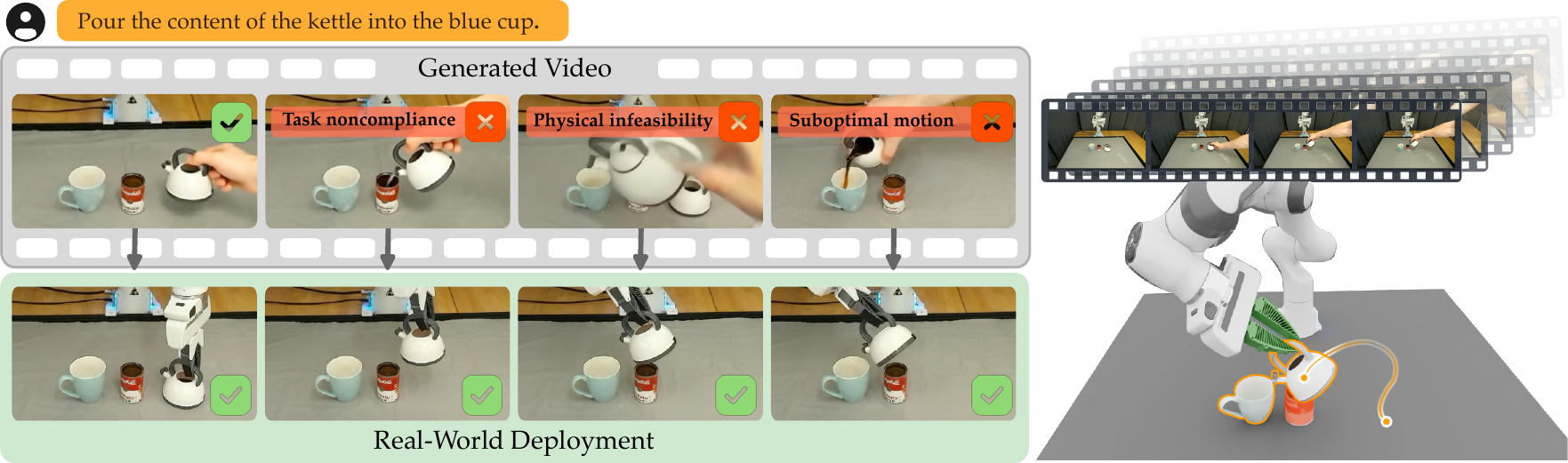}
    \caption{Using video generation models at inference time to guide real-world robot trajectories. The generated video looks plausible, but may make mistakes that prevent directly transferring this video to the real robot's motion. We bridge the gap between generation and deployment by developing an optimization-based formalism. Under this formalism the robot uses multiple synthetic videos to seed its search and identify task-critical features. By matching these features while enforcing semantic consistency and physical feasibility, we can transfer erroneous videos to successful zero-shot trajectories.}
    \label{fig:front}
    \vspace{-1.5em}
\end{figure*}

\input{1_intro}

\input{2_related}
\input{3_problem}
\input{4_unified}
\input{5_method}
\input{6_experiments}

\input{7_conclusion}

\bibliographystyle{IEEEtran}
\bibliography{references}
\input{8_appendix}

\end{document}

%% file: 1_intro.tex
\section{Introduction}\label{sec:intro}

Collecting robot-centric data is expensive \cite{khazatsky2025droid, iyer2024openteach}.
Videos of humans completing tasks are widely available, but these in-the-wild videos are often unstructured, and occur in settings that are unlike the robot's context.
Video generation techniques can help bridge this gap.
By prompting video generators to create \textit{synthetic} videos, we gather ``human'' data without actually requiring human demonstrations --- plus, the resulting videos are set in the robot's current environment.

Consider the example in \fig{front}.
The user prompts the system with a desired task.
Given this prompt and an initial image, recent video generation models can output a synthetic video of the task being performed by a human.
Ideally, the robot would be able to directly apply this video to guide its own motion.
\textit{But there are two key limitations.}
First, while the synthetic video may appear visually plausible, it is not necessarily physically feasible.
Second, the generated video features a human doing the task, introducing kinematic and visual discrepancies between what the video shows and what the robot arm can actually do.

Our work focuses specifically on using pretrained video generation models \textit{at inference time} to guide robot behavior.
That is, rather than using generated videos as offline data for policy training or fine-tuning a video generator on robot-specific data, we consider a setting where a general-purpose, frozen video generator produces task demonstrations during deployment.
Existing research searches for ways to address the above problems and unlock the promise of video generation models in this setting.
Common approaches typically fall along a spectrum in how much they post-process the video data.
At one extreme, the system simply learns to project the object's motion in the synthetic video to real-world robot trajectories \cite{patel2025robotic}.
At the other extreme, robots construct a digital twin to physically test and refine the proposed video behavior before applying the results in the real world \cite{physworld}.
We view these methods as effective \textit{ad hoc} solutions, but it remains unclear \textit{how to define the underlying problem}.
For example: what should the robot optimize for when trying to map the synthetic video to real-world task performance?
And what characteristics does the video generation model need to have in order for this zero-shot transfer to be possible?

In this paper we seek to formalize the process of converting synthetic videos to real-world robot trajectories for zero-shot manipulation tasks.
Our key insight is that:
\begin{center} \vspace{-0.3em}
    \textit{Transferring video generation is a constrained optimization problem. The synthetic video seeds the search, which should maximize trajectory feature alignment while maintaining semantic and feasibility constraints.}
    \vspace{-0.3em}
\end{center}
Building on this formalism, we introduce \textbf{CASTER}, \textbf{C}onstraint-\textbf{A}ware \textbf{S}ynthetic \textbf{T}ask-feature \textbf{E}xt\textbf{R}action, a framework that extracts task-relevant features from generated videos and optimizes robot trajectories within a digital twin.
The extracted trajectories serve as initial conditions: within a digital twin, the robot then fine-tunes those trajectories to ensure that its resulting behavior performs the intended object motions without unintended collisions or physically infeasible behaviors.
We find that the objective function itself is particularly important.
While prior works assume that the robot should match \textit{every} waypoint of the synthetic video, we generalize so that the robot's objective function assigns higher weights to \textit{task-critical} waypoints or features.

Formalizing this optimization provides three fundamental benefits.
First, we can unify existing approaches as instances of our optimization problem under different simplifying assumptions.
Second, we can use the optimization structure to define the characteristics of the video generation that are necessary for success.
Finally, we can develop new methods for zero-shot manipulation using video generation models by more closely approximating the underlying problem.
Overall, we make the following contributions:

\p{Defining a Unified Formalism}
We define a constrained optimization problem that inputs synthetic task videos and outputs trajectories for real-world execution.
This optimization must satisfy physical feasibility (i.e., the motion must be executable without accidental collisions) and semantic consistency (i.e., the motion must be aligned with the task).

\p{Characterizing Video Generation}
Building on our theoretical framework, we derive the properties and requirements for the video generation model.
We show that simultaneously enforcing feasibility and semantic constraints depends on generating videos that can be locally modified without losing their semantic meaning.
Intuitively, tasks with a larger basin of attraction around feasible motions (i.e., many initial guesses converge towards the same equilibrium) are more suitable for synthetic video frameworks.

\p{Developing Approximate Solutions}
We demonstrate that state-of-the-art approaches can be viewed as instances of our unified framework.
We then derive a new approach that more closely matches the underlying optimization problem by weighting trajectory features instead of matching every waypoint.
Our experiments show the pros and cons of these different solution methods.

%% file: 2_related.tex
\section{Related Work} \label{sec:related}

\noindent \textbf{Robot Video Generation for Manipulation.}
Recent works have explored generating videos of robots performing tasks for learning manipulation policies and planning robot actions\cite{ko2023avdc}. 
Generating the robot directly reduces the differences in embodiment between the video demonstration and robot execution\cite{xu2026dream2act}.
UniPi~\cite{du2023learning} fine-tunes a pretrained video generator on robot data to predict language-conditioned visual plans, and uses an inverse-dynamics model to recover robot actions. 
Video Policy~\cite{liang2025video} combines video and action diffusion, conditioning the action network on intermediate video features at each denoising step. VAG~\cite{lang2026vagdualstreamvideoactiongeneration} jointly generates aligned video-action pairs, which can be used as synthetic training data or for learning world-action models.

Other works use generated robot videos to augment the data available for policy learning\cite{wang2026robovip,chen2026craft}.
DreamGen~\cite{jang2025dreamgen} generates synthetic trajectories and recovers pseudo-action labels for training robot policies. 
Ctrl-World~\cite{guo2026ctrlworldcontrollablegenerativeworld} predicts action-conditioned, multi-view robot rollouts and uses successful trajectories to fine-tune policies. 
GE-Sim 2.0~\cite{qiu2026gesim2} additionally predicts proprioceptive states and filters generated experience with a learned judge. 
For humanoid manipulation, GenHOI~\cite{bi2026genhoicontactawarehumanoidobjectinteraction} recovers contact-aware geometric constraints from generated interaction videos and optimizes trajectories for execution without task-specific policy training.

However, generating consistent robot videos zero-shot remains challenging for general-purpose video models, as robot behavior is sparsely represented in their training data. 
\textit{These models are trained on large amounts of human video and are better suited to generating human demonstrations in new scenes without fine-tuning}~\cite{bharadhwaj2024gen2act}. 
We therefore use human video generation to predict how a task can be performed, and translate the generated motion into robot actions.

\p{Human Video Generation for Manipulation}
Generated human videos can describe tasks across diverse objects and scenes, but do not directly specify executable robot actions.
Prior works extract motion from these videos for trajectory optimization or policy learning~\cite{ziakas2026grounding}, including approaches that generate object motion without a visible human.

One line of work recovers trajectories or keyframes and translates them into robot actions through retargeting or trajectory optimization~\cite{paliwal2026do, zhang2026emboalign, hu2026v2strep, huang2026genvid2robot, chen2025lvp}.
Dreamitate~\cite{liang2024dreamitate} tracks tool motion, while RiGVid~\cite{patel2025robotic} recovers object poses; both use these trajectories to determine robot end-effector motion. 
NovaFlow~\cite{li2025novaflow} recovers 3D object flow, and NovaPlan~\cite{fu2026novaplan} uses object and hand motion with closed-loop replanning.
RoboReact~\cite{he2026roboreact} reconstructs interaction keyframes, retargets them to the robot, and refines execution using feedback.

Other works learn policies that use generated videos or recovered motion\cite{kuang2026dex4d}. Gen2Act~\cite{bharadhwaj2024gen2act} trains a closed-loop policy conditioned on generated human videos to perform tasks specified through video generation at deployment. 
GenMimic~\cite{ni2025genmimic} trains a physics-aware reinforcement learning policy to track recovered human motion. 
Dream2Flow~\cite{dharmarajan2025dream2flow} reconstructs 3D object flow and uses a tracking objective for either trajectory optimization or reinforcement learning. 
PhysWorld~\cite{physworld} uses reinforcement learning in a digital twin to obtain executable robot actions that track generated object motions.
We seek a generalized formulation that lets us study these prior approaches within a common framework, analyze when generated motion can successfully transfer to a robot, and explicitly bridge semantic intent and physical feasibility.

%% file: 3_problem.tex
\section{Problem Statement} \label{sec:problem}

We consider the setting of zero-shot tabletop robot manipulation.
An example of this setting is shown in \fig{front}.
The robot is given a natural-language task description $\tau$ and an initial visual observation $o$ of a scene containing $M$ objects. 
The setting is \textit{zero-shot} in the sense that the robot must execute the task within a finite horizon without access to task-specific robot demonstrations (i.e., teleoperation) or any additional real-world interaction.

\p{Trajectories}
We assume a fixed tabletop workspace observed by a static camera where task-relevant objects remain at least partially observable during execution.
Let $s_o$ be the state of the $M$ objects (e.g., the pose of each item) and let $s_r$ be the robot's state (e.g., its joint and gripper positions).
The system state $s = (s_o, s_r)$ includes both the objects and the robot; this state transitions based on robot actions $a$.
In our experiments these actions are changes in joint position of the arm and commands to open or close the gripper.
During an interaction of $T$ timesteps the robot executes actions $a^{0:T-1}$, resulting in a sequence $s^{0:T} = (s_o^{0:T}, s_r^{0:T})$ of system states.
We therefore define the \textit{trajectories}:
\begin{equation} \label{eq:P1}
    \xi = \left(s^{0:T}, a^{0:T-1}\right), \quad \xi_o = s_o^{0:T}, \quad \xi_r = s_r^{0:T}
\end{equation}
where $\xi_o$ and $\xi_r$ isolate the objects and robot motions. 
Similarly, we define $\xi_a = a^{0:T-1}$ as the trajectory of robot actions that induce these object and robot motions.

\p{Task Success}
For an interaction to be successful the system's trajectory $\xi$ must complete the task $\tau$.
This includes both the \textit{semantic} output (e.g., pouring water into the glass) and the \textit{physical} requirements (e.g., avoiding collisions with that glass). 
We express these requirements as:
\begin{equation} \label{eq:P2}
    \mathcal{G}_{\tau}(\xi) = 1,
    \quad
    \xi \in \mathcal{F}_{\mathcal{R}}
\end{equation}
where $\mathcal{G}_{\tau}$ is a task-dependent success predicate and $\mathcal{F}_{\mathcal{R}}$ is the set of trajectories permitted by the robot embodiment and environment.
This feasible set captures constraints such as workspace reachability, joint and velocity limits, collision avoidance, and the ability to establish the contacts required to manipulate relevant objects. 

\p{Video Generation}
In addition to the initial observation $o$ and task description $\tau$, we assume inference-only access to a pretrained image-and-text-conditioned \textit{video generation model}. 
Given $(o,\tau)$, the video generation model $q_{\mathrm{vid}}$ defines a conditional distribution over possible future videos:
\begin{equation} \label{eq:P3}
    v \sim q_{\mathrm{vid}}\!\left(v \mid o,\tau\right), \quad \xi_{o}^{\mathrm{video}} = f_{\mathrm{track}}(v)
\end{equation}
In \eq{P3} $v$ is the raw RGB video and $\xi_{o}^{\mathrm{video}}$ is the motion trajectory of the $M$ objects in that video.
For now we assume access to a procedure $f_{\mathrm{track}}$ that accurately maps the synthetic video into these object trajectories.

The generator $q_{\mathrm{vid}}$ is treated as a \textit{black box}: its parameters, training data, gradients, and internal representations are unavailable, and it cannot be fine-tuned using robot data.
The generated videos do not necessarily depict the target robot or provide any explicit geometric, contact, or kinematic information.
These videos may therefore describe futures that are semantically consistent with the task but infeasible for the robot, or physically plausible motions that do not accomplish the requested task. 
Our goal is to use only the robot’s observations, its embodiment and control model, and samples obtained from this frozen video generator to perform a manipulation task that satisfies both the semantic success and physical feasibility from \eq{P2}.

%% file: 4_unified.tex
\section{Introducing a Unified Formalism} \label{sec:unified_formalism}

Generated videos provide insights into the correct behavior through their visual content, but these videos do not directly provide executable robot actions.
Prior work has explored several approaches for zero-shot robot manipulation based on synthetic video generation.
Rather than starting with another solution, \textit{we begin by focusing on the underlying problem}.
More specifically, we formulate leveraging synthetic videos as constrained optimization.
In Section~\ref{sec:U1} we define the generalized components of this optimization, and in Section~\ref{sec:U2} we derive the properties the video generator must satisfy in order for the optimization to converge.
Finally, in Section~\ref{sec:U3} we show how related approaches can be viewed as instances of our unified formalism with relaxed constraints or different cost functions.

\subsection{\small{Learning from Generated Videos as Constrained Optimization}} \label{sec:U1}

Given a synthetic video $v$ of task $\tau$, our goal is to extract robot actions $\xi_a$ that perform the same task with a real-world robotic system.
When the robot takes these actions, they result in the objects, robot, and system trajectories from \eq{P1}.
Ideally these trajectories produce task success.
Consider \eq{P2}: the behavior resulting from $\xi_a$ must be both semantically aligned and physically feasible.
The synthetically generated video does not directly show what actions the robot should take, but in \eq{P3} it does provide insight into what the object trajectories could be.
Put another way, each synthetic video yields an \textit{initial guess} at the optimal object motions $\xi_{0,v}$.
The motion resulting from $\xi_a$ should be \textit{close} to that initial guess, while satisfying the \textit{task success constraints}.
We write this formally as a constrained optimization problem:
\begin{equation} \label{eq:U1}
\begin{aligned}
& \xi_a^* \in \arg\min_{\xi_a} && \left\| \phi(\xi_o) - \phi\big(f_{\mathrm{track}}(v)\big) \right\| \\
&\text{subject to:} && \mathbb{P}\left[\mathcal{G}_{\tau}(\xi) = 1\right] \geq 1 - \epsilon_{\mathrm{sem}} \\
&&& \mathbb{P}\left[\xi \in \mathcal{F}_\mathcal{R}\right] \geq 1 - \epsilon_{\mathrm{phys}}
\end{aligned}
\end{equation}
where $\xi_o$ and $\xi$ are the trajectories induced by robot actions $\xi_a$.
Below we separately discuss the cost function and constraints from \eq{U1}.

\p{Cost Function}
The optimization cost captures the distance between the object motions in the video and the object motions in the real-world trajectory.
More specifically, we focus on the deviation in \textit{task features}.
Although task features can be instantiated in various ways, we generally represent them as relative trajectories between objects.
Let $\phi : \Xi_o \rightarrow \mathbb{R}^k$ map the $M$ object trajectories into a $k$-length vector.
In our problem setting, trajectory length is constrained by video generation, and we assume that corresponding timesteps across videos represent comparable stages of task execution.
One straightforward choice for $\phi$ is the identity mapping.
Here the features are simply the object poses, and minimizing the cost causes the robot's actions to try and recreate every single object motion in the synthetic video.

In practice, however, some aspects of the object motions are more task-critical than others.
Consider the task ``pour a glass of water.''
For this task, the exact path the robot follows to carry the water to a cup is not particularly important, provided the robot keeps the water upright.
The features $\phi$ therefore only need to capture the \textit{task-critical} parts of the object motions and inter-object geometric relationships.
These features can discard noise and incidental variations in the motions and timing of the generated trajectories.

\p{Semantic Consistency Constraint}
Optimizing for the object features alone is insufficient: even if we faithfully reproduce the object trajectories in the video, these trajectories may not complete the intended task.
We therefore enforce constraints in \eq{U1}.
To understand how these constraints connect to the properties of the video generation model in Section~\ref{sec:U2}, we \textit{expand on their definition here}, starting with the semantic consistency constraint.

Given the task-dependent success predicate $\mathcal{G}_\tau$, we define the corresponding semantic solution manifold for task $\tau$ and initial observation $o$ as:
\begin{equation} \label{eq:U2}
\mathcal{M}_{\mathrm{sem}}(\tau,o) =
\left\{
\xi_a :
\mathcal{G}_{\tau}(\xi)=1
\right\}
\end{equation}
This manifold contains all the action trajectories that achieve the intended task, regardless of their motions and timings.
Because our cost function compares trajectories through the lens of object features, we also define the image of the semantic solution manifold in the feature space:
\begin{equation} \label{eq:U3}
\mathcal{M}_{\mathrm{sem}}^{\phi}(\tau,o) =
\left\{
\phi(\xi_o) : \mathcal{G}_{\tau}(\xi)=1
\right\}
\end{equation}
where we emphasize that the objects' trajectory $\xi_o$ is induced by robot actions $\xi_a$.
In practice, our optimization procedure may not be able to satisfy the semantic constraint with absolute certainty.
We therefore relax the success condition from \eq{P2} to reach a more general constraint:
\begin{equation} \label{eq:U4}
\mathbb{P}
\left[
\mathcal{G}_{\tau}(\xi)=1
\right]
\geq
1-\epsilon_{\mathrm{sem}}
\end{equation}
where $\epsilon_{\mathrm{sem}} \in [0,1)$ is the tolerated probability of semantic failure.
In summary, we constrain the actions $\xi_a$ such that the resulting object and robot trajectories complete the intended task $\tau$ with high probability.

\p{Physical Feasibility Constraint}
The physical feasibility constraint requires the state-action trajectory to remain within the set of behaviors the robot can actually execute.
Similar to our semantic approach, we relax \eq{P2} to define this constraint probabilistically:
\begin{equation} \label{eq:U5}
\mathbb{P}
\left[
\xi \in\mathcal{F}_{\mathcal{R}}
\right]
\geq 1-\epsilon_{\mathrm{phys}}
\end{equation}
where $\epsilon_{\mathrm{phys}} \in [0,1)$ is the tolerated probability of physical constraint violation.
The feasible set $\mathcal{F}_{\mathcal{R}}$ is determined by the robot embodiment, the environment, and the interactions required by the task.
Given task $\tau$ and initial image $o$ let:
\begin{equation} \label{eq:U6}
g_j(\xi;\tau,o)\leq 0,
\qquad j=1,\ldots,J
\end{equation}
denote the individual physical constraints, such that:
\begin{equation} \label{eq:U7}
\mathcal{F}_{\mathcal{R}}
=
\left\{
\xi:
g_j(\xi;\tau,o)\leq 0,\ 
j=1,\ldots,J
\right\}
\end{equation}
is the feasibility set.
In practice, the robot can determine if it satisfies \eq{U5} by checking each of the individual constraints in \eq{U7}.

One standard approach for enforcing physical feasibility is utilizing a \textit{digital twin}.
Let $\hat{g}^{\mathrm{sim}}_j$ denote the simulator's approximation of the real constraint $g_j^{\mathrm{real}}$ from \eq{U6}. 
The policy can then be optimized subject to:
\begin{equation} \label{eq:U8}
\mathbb{P}
\left[
\hat{g}^{\mathrm{sim}}_j(\xi;\tau,o)
\leq -\delta_j,\ 
\forall j
\right]
\geq 1-\epsilon_{\mathrm{phys}}
\end{equation}
where $\delta_j\geq 0$ is a safety margin accounting for mismatch between the
digital twin and the real system.
Intuitively, this safety margin means the robot must satisfy the physical constraint with an \textit{additional tolerance} in the digital twin; increasing this tolerance improves sim-to-real robustness.
If:
\begin{equation} \label{eq:U9}
\left|
g_j^{\mathrm{real}}(\xi;\tau,o)
-
\hat{g}^{\mathrm{sim}}_j(\xi;\tau,o)
\right|
\leq \delta_j
\end{equation}
over the relevant trajectory set, then the added tolerance is sufficient and $\hat{g}^{\mathrm{sim}}_j(\xi;\tau,o)\leq-\delta_j$ implies
$g_j^{\mathrm{real}}(\xi;\tau,o)\leq0$.
Put another way, if \eq{U9} holds then satisfying the stronger constraint in the digital twin will also satisfy the original feasibility constraint from \eq{U5}.

\subsection{Video Generation Requirements}
\label{sec:U2}

Thus far we have framed zero-shot robot manipulation via synthetic video generation as a constrained optimization problem.
Under this framework the video generator provides an \textit{initial guess} at the solution, which is then refined by \eq{U1}.
So what properties must the video generator satisfy?
To answer this question we will define a basin of attraction, i.e., a set of initial guesses that all converge to the same equilibrium set.
More specifically, we focus on \textit{semantic basins of attraction}.
These semantic basins provide local regions of the trajectory manifold that when constrained for physical feasibility converge to motions which are aligned with the task-dependent success predicate $\mathcal{G}_\tau$.
Below we show how the success of the video generator depends on seeding the optimization in \eq{U1} within one of these semantic basins of attraction.

\p{Basin of Attraction}
Recall from \eq{P3} that $\xi_{o}^{\mathrm{video}}$ is the motion trajectory of objects in the synthetic video.
We define $z$ as the features extracted from these trajectories:
\begin{equation} \label{eq:U10}
    z = \phi\left(f_{\mathrm{track}}\left(v\right)\right)
\end{equation}
We next define the basin of attraction around these features.
Intuitively, the generated features must be close to both semantic consistency and physical feasibility:
\begin{equation} \label{eq:U11}
\mathcal{S}_{\rho}^{\phi}(\tau,o)
=
\left\{
z:
d_{\phi}\!\left(
z,
\mathcal{M}_{\mathrm{sem}}^{\phi}(\tau,o)
\right)
\leq
\rho(\tau,o)
\right\}
\end{equation}
Here $\mathcal{M}_{\mathrm{sem}}^{\phi}$ from \eq{U3} is the set of object trajectories that perform the intended task, $z$ from \eq{U10} is the generated features, and $d_\phi$ is the L2 distance in feature space.
\eq{U11} states that this distance must be less than or equal to the radius $\rho(\tau, o) \geq 0$.
The radius characterizes how far the features of a video-derived trajectory may deviate from the semantic feature manifold while still allowing the given physical-constraint enforcement mechanism to preserve its semantic mode.
This radius may depend on the task, the initial scene, the robot and environment, and the properties of the enforcement mechanism.

To summarize: we assume that \eq{U11} defines a sufficient basin of attraction for the chosen optimization procedure.
Under this assumption, synthetic videos which yield object trajectories $\xi_{o}^{\mathrm{video}}$ and features $z$ within $\mathcal{S}_{\rho}^{\phi}(\tau,o)$ converge to trajectories that successfully complete the task.

\p{Video Generation Bound}
The probability that the synthetic video lies within the relevant basin of attraction depends on the quality of the video generation model.
Because the generated videos $v$ are stochastic, the extracted feature values $z$ can be viewed as a random variable.
Let $z_{\mathrm{sem}} \in
\mathcal{M}_{\mathrm{sem}}^{\phi}(\tau,o)$ be a semantically valid feature realization.
Write
\begin{equation} \label{eq:U12}
z = z_{\mathrm{sem}} + \eta
\end{equation}
where $\eta$ denotes the error in the desired feature values extracted from the generated videos.
By definition, the expected squared feature error is:
\begin{equation} \label{eq:U13}
\mathbb{E}\left[\|\eta\|_2^2 \mid \tau,o \right]
=
\sigma_{\mathrm{vid}}^2(\tau,o)
\end{equation}
For $\rho(\tau,o)>0$ and finite $\sigma_{\mathrm{vid}}^2(\tau,o)$, Markov's inequality therefore gives
a probabilistic condition on video generation:
\begin{equation} \label{eq:U14}
\mathbb{P}
\left[
z\in
\mathcal{S}_{\rho}^{\phi}(\tau,o)
\mid
\tau,o
\right]
\geq
1-
\frac{
\sigma_{\mathrm{vid}}^2(\tau,o)
}{
\rho^2(\tau,o)
}
\end{equation}
Put intuitively: the probability of successfully seeding the optimization within a basin of attraction depends on the ratio between (a) the feature error in the synthetic video, $\sigma_{\mathrm{vid}}$, and (b) the radius within which the physical constraint can be successfully applied, $\rho$.
In the best-case scenario (\textit{reducing} $\sigma_{\mathrm{vid}}$) the generated video moves the objects to complete the commanded task, and (\textit{increasing} $\rho$) the object trajectories can be locally modified for physical feasibility without losing their semantic meaning. 
This analysis does not require unbiased features. Systematic
hallucinations from the video model are allowed, but can weaken
the bound by increasing $\sigma_{\mathrm{vid}}^2$.

\p{Task-Selection Implications}
The sufficient condition outlined by \eq{U14} offers insights into which types of tasks are suitable for our synthetic video pipeline.
\textit{The more the object trajectories can be modified without violating their semantic meaning, the better suited for video generation}.
Accordingly, pick-and-place tasks are favorable, since the exact motion used to go from the pick to the place can be highly variable (increasing $\rho$).
At the other extreme, contact-rich tasks like pushing an object into a crowded box are unfavorable, since small changes in pose, timing, or grasp can alter the entire task (decreasing $\rho$).

\subsection{Unifying Related Approaches} \label{sec:U3}

In addition to providing guidance on video generation properties, our unified optimization framework also lets us connect recent methods.
We can view the state-of-the-art methods \cite{patel2025robotic, physworld, li2025novaflow} as \textit{ad hoc} instances of \eq{U1} under simplifying assumptions:

\p{PhysWorld~\cite{physworld}} 
PhysWorld extracts a single target object trajectory from a synthetic video, and then optimizes the robot's behavior in a digital twin to track this trajectory.
This is equivalent to \eq{U1} but with a more restrictive cost function:
\begin{equation} \label{eq:U15}
    \xi_a^* \in \arg\min_{\xi_a} \left\| \xi_{\mathrm{target}} - \xi_{\mathrm{target}}^{\mathrm{video}} \right\|
\end{equation}
where $\xi_{\mathrm{target}}$ is the trajectory of the target object. Related approaches such as \cite{li2025novaflow} map generated rigid object motions recovered from a single video to end-effector poses, then optimize trajectories with collision and joint limit penalties.

\p{RIGVid~\cite{patel2025robotic}} 
RIGVid directly replays the object trajectory extracted from the synthetic video by assuming a rigid grasp, and using kinematics to map from object motion to robot motion.
No physical feasibility is explicitly enforced.
This is equivalent to \eq{U1} with two changes: (a) the more specific tracking cost from \eq{U15} and (b) no physical feasibility constraint $\mathbb{P}\left[\xi \in \mathcal{F}_\mathcal{R}\right] \geq 1 - \epsilon_{\mathrm{phys}}$.

\smallskip

We note that these simplifications are not necessarily a negative; they may be reasonable choices for some problem settings.
We will explore how these simplifications compare to more closely following \eq{U1} in our experiments.

%% file: 5_method.tex
\begin{figure*}[t]
    \centering
    \includegraphics[width=\textwidth]{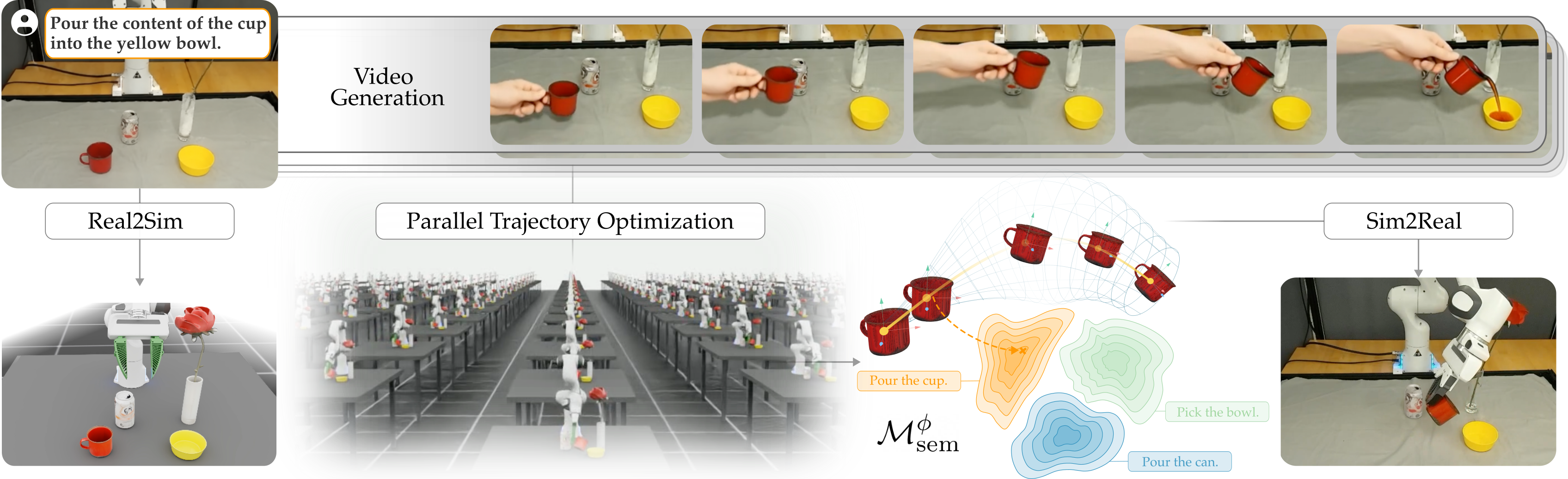}
    \caption{Overview of \textbf{CASTER}. (Top Left) Given an initial RGB-D observation and a task prompt, we (Top) generate synthetic videos of a human performing that task in the current environment and (Bottom Left) build a digital twin of the scene. From the synthetic videos we extract the object trajectories, which then serve as initial guesses for trajectory optimization within the digital twin. (Bottom) The trajectories extracted from synthetic videos do not need to be perfect: instead, they must lie within a semantic basin of attraction where the optimization can locally modify the trajectory shape without removing its semantic meaning. (Bottom Right) We perform constrained optimization so that the robot's actions in the digital twin cause the objects to move similarly to their video trajectories, while enforcing kinematic constraints so that the output can be directly executed by the real robot.}
    \label{fig:method}
    \vspace{-1.5em}
\end{figure*}

\section{Approximating the Underlying Solution} \label{sec:method}

In Section~\ref{sec:unified_formalism} we present zero-shot manipulation from synthetic videos as a constrained optimization problem.
While current methods introduce simplifications, in this section we present \textbf{CASTER} (Figure~\ref{fig:method}), an approach that more closely follows \eq{U1}.
Practically, this leads to two distinctions from 
previous work.
First, instead of optimizing to exactly follow the object's video trajectory, we preserve \textit{task features} $\phi$.
Second, we consider multiple sampled synthetic videos when extracting these features, enabling the robot to optimize over \textit{expected behaviors}.
We build a digital twin to reconstruct the scene,
then perform constrained trajectory optimization in the digital twin while enforcing collision-avoidance and kinematic constraints.
The aggregated trajectory features define a shared optimization target and each sampled video initializes a separate optimization run.
The resulting robot trajectory can be executed open-loop in the real world.

\subsection{Extracting the Task Features} \label{sec:M1}

We start with the cost function in \eq{U1}.
Using the task instruction $\tau$ and a single RGB-D observation $o = [\mathbf{I}, \mathbf{d}] \in \mathbb{R}^{H \times W \times 4}$, the robot applies \eq{P3} to sample multiple videos $v$ from a pretrained video generator.
Each video is then checked by a VLM to validate whether the behavior shown in that video performs the requested task.
Videos which fail this check are discarded, and only the retained videos are used to extract task features.

\p{Tracking Objects} 
Referring to \eq{P3}, we next need to instantiate $f_{\mathrm{track}}$ and extract the object trajectories $\xi_o^\mathrm{video}$ from the synthetic videos.
We assume the planar mapping from the RGB image to the robot's coordinate frame is calibrated.
We then use an off-the-shelf depth generation model \cite{depthanything3} to estimate depth for each frame of the generated video, denoted as $\hat{\mathbf{d}}_{1:T}$.
To align these estimates with the observed depth $\mathbf{d}$, we fit an affine transformation to the estimated depth of the first frame $o$ and apply the same transformation to all subsequent frames.
To compute the object orientation in the synthetic videos we apply TrackCraft3R \cite{nam2026trackcraft3} --- this approach tracks a dense set of 3D points on the object surface, which can then be fit to a rigid body transformation.
Put together, this position, depth, and orientation tracking produces $\xi_{o}^{\mathrm{video}}$.

\p{Extracting Features}
In Section~\ref{sec:U1} we argue that the robot should not try to \textit{exactly reproduce} every object trajectory.
Instead, the robot only needs to match the task-critical parts of the object motions in $\xi_{o}^{\mathrm{video}}$.
To identify these critical features we prompt an LLM with the semantic task $\tau$ and the object labels (e.g. the names of each object in the scene).
This LLM outputs two things: (a) the \textit{object relations} essential to the task and (b) the \textit{trajectory dimensions} that need to be tracked.
Object relations are defined as target-context object pairs $\xi_{\tau} =\{(\xi_t^1, \xi_c^1), \cdots (\xi_t^K, \xi_c^K)\}$, where $\xi_t, \xi_c \in \xi_o$.
For example, when ``pouring the coffee'' a target could be the coffee mug and the context could be the coffee pot.
The trajectory dimensions refer to the specific elements of these motions which should be tracked (e.g., tracking the position specifically in the $x$ and $y$ axes).

The resulting LLM outputs are then applied to each video $v$ that we have sampled for the task.
We compute $\bar{\xi}_\tau$, the \textit{median position and mean orientation} for the object relations along dimensions identified by the LLM.
We also consider the \textit{variance} among these sampled videos.
Specifically, we compute time-dependent \textit{weights} $w$ which decay exponentially with cross-demonstration variance and emphasize consistent features across all the sampled videos.
This overall process of using an LLM to identify relevant object relations and then calculating their mean and variance across multiple synthetic videos constitutes the feature extractor $\phi$.

\subsection{Constrained Optimization in a Digital Twin} \label{sec:M2} 

We optimize \eq{U1} in a digital twin.
Within our experiments we use Isaac Sim \cite{NVIDIA_Isaac_Sim}, and reconstruct objects from the initial observation $o$ using SAM 3D \cite{chen2026sam}.
Optimization is performed over the robot's end-effector trajectory \textit{splines} with control parameters $\xi_a \in \mathbb{R}^{6P}$.
Here $P$ is the number of control points (selected by the designer), and each point is parameterized as a six-dimensional pose.
We optimize these parameters following the cross-entropy method (CEM)~\cite{CEM}, evaluating each candidate through a simulated rollout with the feature-based cost function:
\begin{equation} \label{eq:M1}
\begin{aligned}
& \xi_a^* \in \arg\min_{\xi_a} \sum_t w_t \cdot \left\| \phi\big(\xi_o^t\big) - \bar{\xi}_\tau^t \right\| \\
&\text{subject to:~} \left\|F(\xi_a)\right\| \leq \bar{F}, \quad q_{\min} \leq q(\xi_a) \leq q_{\max}
\end{aligned}
\end{equation}
Vector $F$ is the force applied by the simulated robot, and $\bar{F}$ is the maximum force threshold when interacting with task-irrelevant objects.
The constraints prevent unintended collisions that exceed $\bar{F}$, as well as joint values $q$ that exceed the robot's kinematic limits.
Comparing our approximation in \eq{M1} to the idealized optimization in \eq{U1}, we notice one key distinction.
\eq{M1} does not directly include the semantic consistency constraint: this is because we pre-process the synthetic videos used to compute the reference motions $\bar{\xi}_\tau$, and ensure these videos perform the intended task.
Based on our analysis from Section~\ref{sec:U2}, if these videos seed the optimization within a semantic basin of attraction their task consistency will be preserved.

%% file: 6_experiments.tex
\section{Experiments} \label{sec:experiments}

\begin{figure}[t]
    \vspace*{6pt}
    \centering
    \includegraphics[width=1\columnwidth]{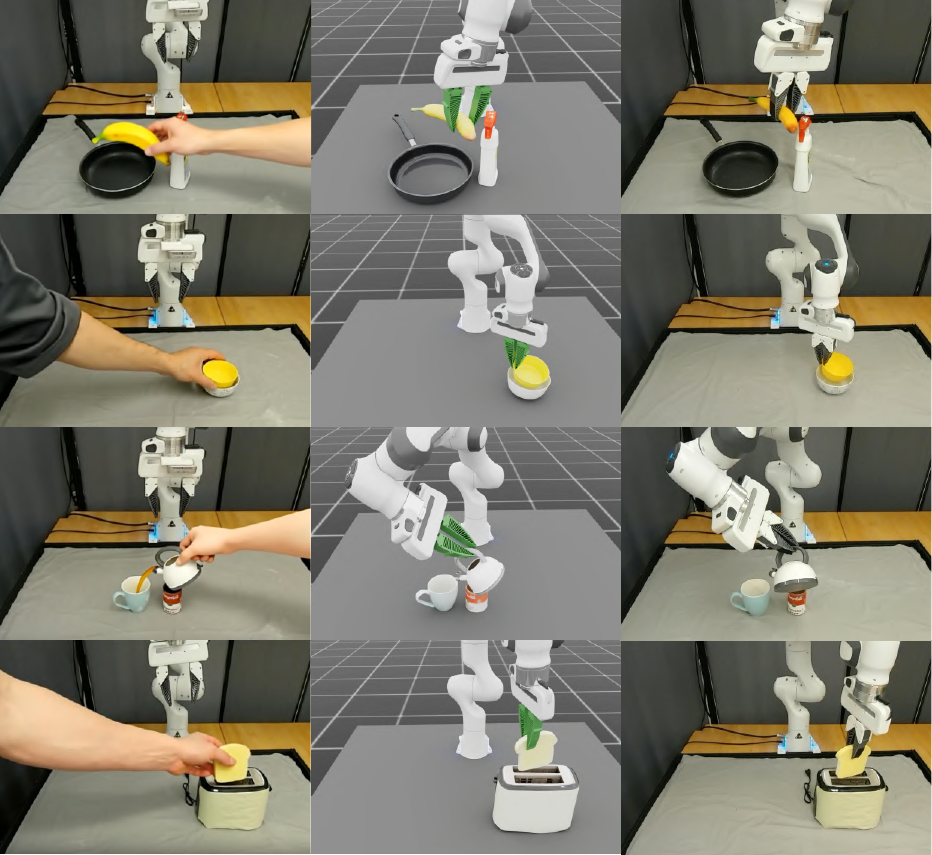}
    \caption{Examples of our real-world tasks. (Left) Generated videos. (Middle) Digital twin rollouts. (Right) Real-world execution. The rows show four tasks: placing a banana in a pan, stacking bowls, pouring from a kettle, and toasting bread. See our supplemental video for more examples.}
    \label{fig:task}
    \vspace{-1.5em}
\end{figure}

We conduct real-world experiments to evaluate the effectiveness of our unified framework.
Within these experiments the robot is expected to perform the 12 tasks from Table~\ref{tab:results} on its first attempt, without any real-world demonstrations or additional interactions.
To find the correct trajectory the robot uses synthetically generated videos of the human performing the task.
Examples of these videos are shown in \fig{task}.
We chose these tasks because they require compliance with semantic instructions and multiple constraints on position, orientation, and collision avoidance.

\p{Baselines}
We compare 
\textbf{CASTER} with two baselines representing common strategies for transferring generated motion, as exemplified by two prior works from Section~\ref{sec:U3}:

\subsubsection{\textup{\textbf{Retarget}}} Similar to RIGVid~\cite{patel2025robotic}, this baseline directly retargets the object trajectory from the synthetic video to the robot's end effector using a rigid-body transformation.

\subsubsection{\textup{\textbf{Mimic}}} Similar to PhysWorld~\cite{physworld}, this baseline optimizes the robot's trajectory in a digital twin so that the resulting object motions precisely mimic the synthetic videos while maintaining physical feasibility.

\p{Setup and Video Generation}
Our real-world setup consists of a Franka Emika robot arm and a stationary Orbbec Femto Mega camera.
The digital twin is constructed on the fly from an initial RGB-D image of the scene using SAM 3D~\cite{chen2026sam} 
and Isaac Sim \cite{NVIDIA_Isaac_Sim}. 
Synthetic videos are generated using HunyuanVideo-1.5 480P-I2V Step-Distilled~\cite{hunyuanvideo2025}.
We generated 20 videos per task using random seeds 0 to 19; all methods were evaluated on the same set of videos. 
Each video is conditioned on the initial scene image and the task-specific text prompts.
The videos contained 121 frames at 24 fps and 480p resolution. 
Prompt rewriting and super-resolution were disabled.

\begin{table}[t]
    \vspace*{6pt}
    \centering
    \small
    \setlength{\tabcolsep}{0.5em}
    \renewcommand{\arraystretch}{1.15}
    \caption{Zero-shot success (\%) when performing real-world manipulation tasks. Mimic and \textbf{CASTER} report mean $\pm$ standard error across seven video initialized CEM runs per task; Retarget reports success rate only.}
    \label{tab:results}
    \begin{tabularx}{\columnwidth}{
        >{\raggedright\arraybackslash}X
        @{\hspace{0.6em}} c c c
    }
        \toprule
        \textbf{Task}
        & \textbf{Retarget} & \textbf{Mimic} & \textbf{CASTER} \\
        \midrule
        1. Move banana in pan
        & $0.0$ & $66.7 \pm 14.5$ & $100.0 \pm 0.0$ \\
        \rowcolor[gray]{0.95}
        2. Pour cup
        & $42.9$ & $52.4 \pm 16.0$ & $95.2 \pm 4.8$ \\
        3. Place block in basket
        & $28.6$ & $76.2 \pm 14.0$ & $100.0 \pm 0.0$ \\
        \rowcolor[gray]{0.95}
        4. Pour kettle
        & $57.1$ & $66.7 \pm 12.6$ & $61.9 \pm 8.7$ \\
        5. Plate orange
        & $57.1$ & $61.9 \pm 15.3$ & $100.0 \pm 0.0$ \\
        \rowcolor[gray]{0.95}
        6. Pile up blocks
        & $0.0$ & $9.5 \pm 6.1$ & $57.1 \pm 20.2$ \\
        7. Stack bowl
        & $14.3$ & $28.6 \pm 15.3$ & $100.0 \pm 0.0$ \\
        \rowcolor[gray]{0.95}
        8. Cover bucket
        & $42.9$ & $28.6 \pm 11.3$ & $33.3 \pm 7.3$ \\
        9. Close lid
        & $14.3$ & $4.8 \pm 4.8$ & $57.1 \pm 14.0$ \\
        \rowcolor[gray]{0.95}
        10. Pick w/o collision
        & $85.7$ & $90.5 \pm 6.1$ & $95.2 \pm 4.8$ \\
        11. Toast bread
        & $28.6$ & $14.3 \pm 9.9$ & $90.5 \pm 6.1$ \\
        \rowcolor[gray]{0.95}
        12. Draw on whiteboard
        & $14.3$ & $9.5 \pm 9.5$ & $28.6 \pm 18.4$ \\
        \midrule
        \textbf{Total}
        & $\mathbf{32.1}$
        & $\mathbf{42.5 \pm 4.5}$
        & $\mathbf{76.6 \pm 3.9}$ \\
        \bottomrule
    \end{tabularx}
    \vspace{-1.5em}
\end{table}

\p{Metrics and Trajectory Sampling}
Task success was defined as achieving the goal specified in the prompt while avoiding collisions with irrelevant objects.
The task failed if the robot tried to violate its joint limits.

After generating the 20 synthetic videos as described above, we next used Qwen3.5 \cite{qwen35blog} as a validator to score the semantic consistency of each video (i.e., did the synthetic video appear to complete the task).
Videos with a score of $\geq 0.75$ were retained.
All the retained videos were used for extracting task features.
We then randomly sampled a set of 7 valid videos for real-world evaluation.
Each of the 7 sampled videos initializes a separate CEM run.
\textbf{Mimic} tracks the corresponding video's object trajectory, whereas \textbf{CASTER} uses the shared feature reference $\bar \xi_\tau$ and weights $w_t$ from all retained videos. For both methods, we discard candidates from each run’s final CEM population that violate collision or joint-limit constraints, then randomly select 3 trajectories from the lowest-cost quartile of feasible candidates. 
Hence, each task comprises seven evaluation trials for \textbf{Retarget} and 21 for \textbf{Mimic} and \textbf{CASTER}. Trials that cannot be executed because of insufficient valid videos or eligible feasible trajectories are counted as failures. For \textbf{Mimic} and \textbf{CASTER}, we report the mean and standard error across the success rates of 7 optimization runs. 


\p{Task Success Results}
Our results are summarized in Table~\ref{tab:results}.
Across the 12 tasks, \textbf{CASTER} achieved a higher average zero-shot success rate than the baselines.
When comparing \textbf{CASTER} and \textbf{Retarget}, we see the advantage of using a digital twin to enforce the physical feasibility constraint.
Since \textbf{Retarget} directly maps the video motion to the robot's trajectory, any infeasible behaviors in the video will carry over to the real world. The resulting \textbf{Retarget} motion frequently collides with surrounding objects.
When comparing \textbf{CASTER} and \textbf{Mimic}, we observe the effect of our feature based, multi-video objective function.
In \textbf{Mimic} the robot is trying to match each waypoint 
and frequently fails to find a solution that satisfies the physical constraints. In contrast, our features only encourage the robot to match waypoints that are consistent across multiple synthetic videos, granting the optimizer greater flexibility.

A major source of failure was the discrepancy between simulated and real-world physics, as illustrated by the ``pour kettle'' and ``cover bucket'' tasks. 
In the former, squeezing the kettle handle causes the kettle to rotate within the gripper, a behavior not captured in simulation. 
In the latter, the bucket's handle has rotational joints that are not represented by our rigid body mesh reconstruction. 
These digital twin discrepancies contribute to the limited performance of both our method and the baselines on these tasks.

\begin{table}[t]
    \vspace*{6pt}
    \centering
    \caption{Evaluating \textbf{CASTER} with different video generation models. Valid reports the number of task-aligned videos out of 80 sampled videos across four tasks. Success reports mean task success (\%) across seven sampled videos per task.}
    \label{tab:video_model_comparison}

    \scriptsize
    \setlength{\tabcolsep}{11pt}
    \renewcommand{\arraystretch}{1.2}

    \resizebox{\columnwidth}{!}{
    \begin{tabular}{l c c}
        \toprule
        \textbf{Model} 
        & \textbf{Valid}
        & \textbf{Success} \\
        \midrule
        
        \textbf{HunyuanVideo-1.5}
        ~\cite{hunyuanvideo2025}
        & $55/80$ 
        & $79.8 \pm 6.5$ \\

        \rowcolor[gray]{0.925}
        \shortstack[l]{
            \textbf{Wan 2.2 I2V-A14B}}
        ~\cite{wan2025}
        & $71/80$ 
        & $76.2 \pm 6.1$ \\

        \shortstack[l]{
            \textbf{LTX-2.3 22B Distilled}
            ~\cite{hacohen2025ltx2}}
        & $52/80$ 
        & $47.6 \pm 8.6$ \\
        
        \rowcolor[gray]{0.925}
        \shortstack[l]{
            \textbf{CogVideoX1.5 5B-I2V}
        ~\cite{yang2024cogvideox}
            }
        & $1/80$ 
        & $0.0 \pm 0.0$ \\

        \bottomrule
    \end{tabular}%
    }
    \vspace{-1.5em}
\end{table}

\p{Video Generator Comparison}
Thus far we have compared our unified approach to state-of-the-art baselines.
But it is possible that these results depend on the specific video generation model used to seed the search.
Accordingly, we here compare four open-source image-to-video models, and see how they impact our method's performance.
The selected models include \cite{hunyuanvideo2025}, \cite{wan2025}, \cite{hacohen2025ltx2}, and \cite{yang2024cogvideox}.
All models receive the same text prompts and initial scene images. 
We use model-specific inference settings and standardize videos to 121 frames at 24 fps before downstream processing.

Our averaged results across tasks 1, 4, 6, and 7 are shown in Table~\ref{tab:video_model_comparison}.
We report the number of synthesized videos that pass the VLM check 
out of 80 videos generated across four tasks, 20 videos per task. 
Overall, we find that validity count alone does not explain downstream success. Wan 2.2 \cite{wan2025} generated more valid videos than HunyuanVideo-1.5 \cite{hunyuanvideo2025}, but achieved similar success, while LTX-2.3 \cite{hacohen2025ltx2} produced a similar validity count yet substantially lower success. We hypothesize that additional valid videos may depict redundant motion strategies that provide limited new guidance. Section~\ref{sec:U2} also suggests that performance may instead depend on task-relevant feature error relative to the optimizer's semantic basin, whose extent depends on task difficulty and robot constraints.

%% file: 7_conclusion.tex
\section{Conclusion} \label{sec:conclusion}

In this paper we 
leverage video generation models at inference time 
for zero-shot manipulation.
By unifying recent works, we introduce a mathematical formalism for transferring the motions 
in generated videos to real robot trajectories.
We pose the setting as a constrained optimization problem.
Using a digital twin, the robot attempts to mimic task-critical aspects of the object trajectories while enforcing 
semantic and physical constraints.
This unified formalism enables
(a) 
establishing a conditional relationship between video feature error and successful initialization,
(b) casting state-of-the-art approaches as instances of our optimization framework, and (c) removing simplifying assumptions from these 
approaches to more closely mirror the formalism.
Across 12 real-world tasks, \textbf{CASTER} achieves 76.6\% average zero-shot task success, exceeding the best-performing baseline by 34.1\%. Future work could train closed-loop policies on \textbf{CASTER} generated trajectories.

%% file: 8_appendix.tex
\appendices
\section{Implementation Details}
\label{app:implementation}

\subsection{Video Generation}

\begin{table}[t]
\centering
\caption{HunyuanVideo inference settings.}
\label{tab:video_specs}
\small
\renewcommand{\arraystretch}{1.2}
\setlength{\tabcolsep}{0.5em}
\begin{tabularx}{\columnwidth}{@{}l>{\raggedright\arraybackslash}X@{}}
\hline
\textbf{Setting} & \textbf{Value} \\
\hline
Sampling steps & 12 \\
\rowcolor[gray]{0.95}
Guidance scale & 1.0 \\
Flow shift & 7.0 \\
\rowcolor[gray]{0.95}
Scheduler & Native flow matching, Euler solver \\
Resolution & 480p, native image-aspect-ratio buckets \\
\rowcolor[gray]{0.95}
Video length / frame rate & 121 frames / 24 fps \\
Transformer precision & BF16 \\
\rowcolor[gray]{0.95}
Prompt rewriting / super-resolution & Disabled / disabled \\
SageAttention / sparse attention & Enabled / disabled \\
\rowcolor[gray]{0.95}
Inference cache & Disabled \\
Inference GPUs & 1 \\
\hline
\end{tabularx}
\vspace{-10pt}
\end{table}

\begin{table}[t]
\centering
\caption{Simulation parameters.}
\label{tab:simulation_specs}
\small
\renewcommand{\arraystretch}{1.2}
\setlength{\tabcolsep}{0.5em}
\begin{tabularx}{\columnwidth}{@{}l>{\raggedright\arraybackslash}X@{}}
\hline
\textbf{Setting} & \textbf{Value} \\
\hline
Physics / control frequency & 120 Hz / 60 Hz \\
\rowcolor[gray]{0.95}
Physics steps per control command & 2 \\
Parallel environments & 1050 \\
\rowcolor[gray]{0.95}
Object mass & 0.08 kg per object \\
Static / dynamic object friction & 4.0 / 3.0 \\
\rowcolor[gray]{0.95}
Object restitution & 0 \\
Gripper stiffness / damping & 4000 / 200 \\
\hline
\end{tabularx}
\vspace{-10pt}
\end{table}

We generate demonstrations using the official HunyuanVideo-1.5 implementation with the \texttt{480p\_i2v\_step\_distilled} checkpoint (Table~\ref{tab:video_specs}). Each video is conditioned on the initial scene image and a task-specific text prompt. We sample 20 videos per task using seeds 0--19, without fine-tuning the generator. The reported experiments use the native \href{https://github.com/Tencent-Hunyuan/HunyuanVideo-1.5}{HunyuanVideo-1.5 Pipeline} rather than the Hugging Face Diffusers implementation. The same checkpoint and seed do not guarantee identical outputs across these implementations. CPU offloading is enabled to reduce GPU memory requirements.

\subsection{Simulation Setup}

We construct a digital twin in Isaac Sim 6.0.1 using Isaac Lab
(Table~\ref{tab:simulation_specs}). Cartesian poses are
mapped to joint targets using damped least-squares differential inverse kinematics.

\subsection{Cross-Entropy Optimization}
\label{app:implementation_C}

We parameterize end-effector position and relative rotation vectors using natural cubic splines with uniformly spaced knots. Orientation is reconstructed as
\begin{equation}
    R(t) = \operatorname{Rot}(\mathbf{r}(t))R_0
\end{equation}
where $R_0$ is the post-grasp orientation and $\operatorname{Rot}(\mathbf{r}(t))$ converts the spline-interpolated rotation vector into a rotation matrix. Its direction gives the
rotation axis, and its magnitude gives the angle in radians. The first rotation knot is fixed at zero, preserving the initial post-grasp orientation. The first position and rotation knots are fixed, giving $D=3(K_{\mathrm{pos}}-1)+3(K_{\mathrm{rot}}-1)$ free parameters. The original task configurations use $(K_{\mathrm{pos}},K_{\mathrm{rot}})=(15,5)$ for all tasks.

The initial mean is obtained by fitting the retargeted demonstration. 
The feature objective constrains only the position and orientation dimensions selected by the LLM feature proposer. For feature objectives that omit certain dimensions, those dimensions are initialized at the post-grasp pose but remain free to optimize. Both Mimic and CASTER use a population of 1024 candidates for up to 10 generations, with initial per-coordinate standard deviations of 0.08 m for position and 0.10 rad for rotation.

At generation $g$, spline parameter vector candidates $\mathbf{x}_g$ are sampled and clipped to the
parameter bounds:
\begin{equation}
    \mathbf{x}^{(i)}_g =
    \operatorname{clip}\!\left(
    \boldsymbol{\mu}_g+
    \boldsymbol{\sigma}_g\odot\boldsymbol{\epsilon}^{(i)},
    \mathbf{l},\mathbf{u}\right),
    \qquad
    \boldsymbol{\epsilon}^{(i)}\sim\mathcal{N}(\mathbf{0},I)
\end{equation}
where $\boldsymbol{\mu}_g$ and $\boldsymbol{\sigma}_g$ are the sampling mean and per-parameter standard deviation at generation $g$, $\mathbf{l}$ and $\mathbf{u}$ are the lower and upper parameter bounds. Position offsets and relative rotation-vector components are bounded by $\pm0.40$ m and $\pm\pi$ rad, respectively, with bounds expanded when necessary to contain the fitted initialization. The lowest-cost 10\% of candidates (102 samples) form the elite set $\mathcal{E}_g$. We update the sampling distribution using the per-parameter elite mean and standard deviation, adding $10^{-4}$ to prevent the standard deviation from collapsing to zero:
\begin{equation}
\begin{aligned}
    \boldsymbol{\mu}_{g+1}
        &= \operatorname{mean}(\mathcal{E}_g), \\
    \boldsymbol{\sigma}_{g+1}
        &= \operatorname{std}(\mathcal{E}_g)+10^{-4}
\end{aligned}
\end{equation}

The best candidate found so far is retained in each population. Candidates are evaluated through parallel physics rollouts. Mimic scores absolute object trajectory agreement, whereas CASTER scores
the shared task feature reference. For the real-world evaluation, we filter the final population for feasibility and randomly select three trajectories from its lowest-cost quartile, as described in the main paper.

\section{Additional Results}

Figure~\ref{fig:additional_overview} summarizes the number of sampled videos that pass validation and the number of feasible trajectories produced in the final CEM generation for each task. Video validation rates vary substantially across tasks, but an average of 70.4\% of videos among the 20 generated videos per task pass the VLM validator, suggesting that scaling up video generation can yield a substantial pool of validated videos. Validation rates are particularly low for tasks requiring precise semantic alignment, such as “pile up blocks” and “toast bread,” which is as expected.

Using the optimization settings described in Appendix~\ref{app:implementation_C}, we compare the number of feasible trajectories produced by \textbf{Mimic} and \textbf{CASTER} from the same demonstrations for each task. We consider a trajectory feasible if it is collision free and remains within the robot’s workspace limits. This criterion does not assess whether the trajectory successfully completes the task. Because the number of validated videos varies across tasks, we report the fraction of feasible candidates per demonstration rather than total candidate counts. We observe that \textbf{CASTER} consistently produces a higher fraction of feasible trajectories on most tasks. This advantage likely stems from tracking only waypoints that preserve key task features, allowing greater flexibility than \textbf{Mimic}, which tracks the entire generated trajectory. In addition to reporting the number of feasible trajectories, Figure~\ref{fig:trajectory} visualizes their spatial distributions for three representative tasks. These examples suggest that \textbf{CASTER} produces more diverse feasible trajectories than \textbf{Mimic}, exploring a broader range of motions while preserving key task features.

\begin{strip}
\centering
\begin{minipage}{\textwidth}
    \centering
    \includegraphics[
        width=\linewidth,
        height=0.22\textheight,
        keepaspectratio
    ]{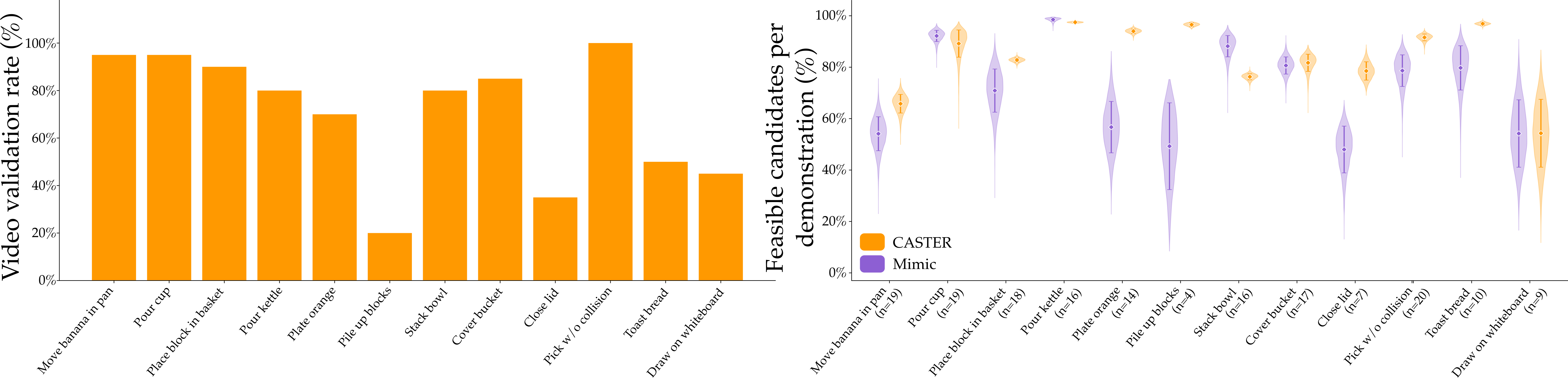}
    \captionof{figure}{
        Left: fraction of the 20 videos generated per task with
        HunyuanVideo-1.5~\cite{hunyuanvideo2025} that pass the VLM
        validator.
        Right: feasible candidate fractions for CASTER (orange)
        and Mimic (purple), using matched demonstrations within
        each task. Diamonds indicate the mean across demonstrations,
        and error bars show one standard error. Violins show Bayesian
        bootstrap posteriors of task means from 10,000 random
        reweightings of matched demonstrations.
        $n$ denotes the number of matched demonstrations.
    }
    \label{fig:additional_overview}

    \vspace{8pt}

    \includegraphics[
        width=\linewidth,
        height=0.22\textheight,
        keepaspectratio
    ]{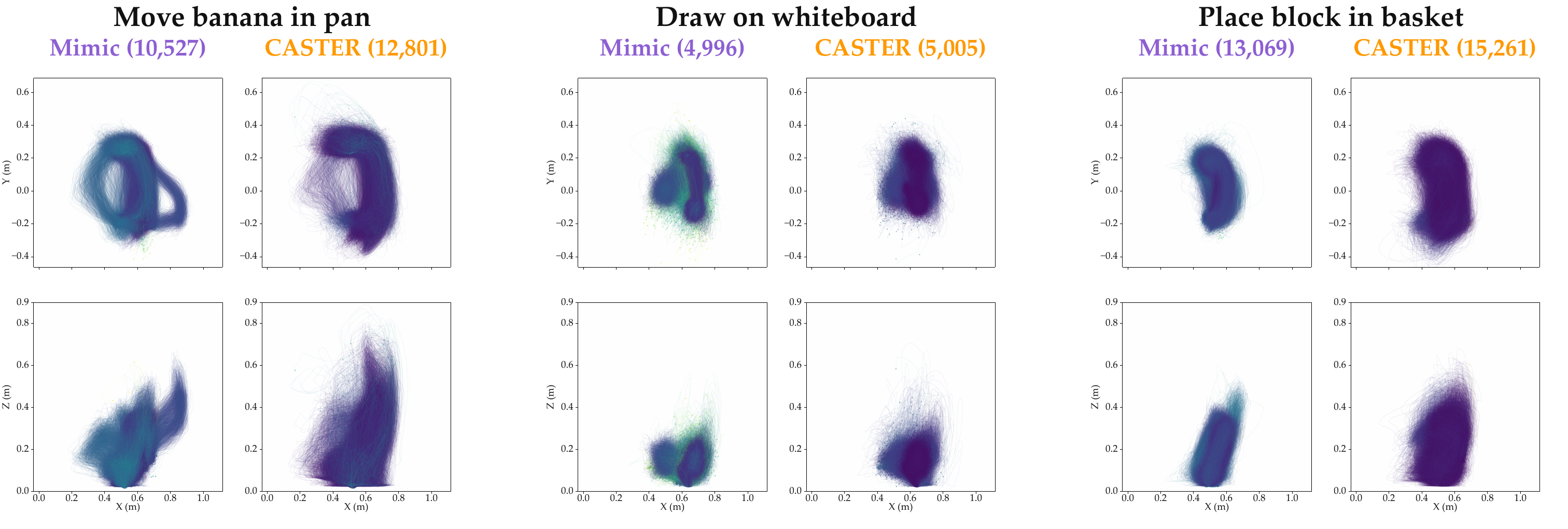}
    \captionof{figure}{
        End-effector trajectories from validated videos produced by
        CASTER and Mimic over the same set of 20 generated videos
        per task, shown in the X--Y (top) and X--Z (bottom) planes
        for three manipulation tasks. Parentheses indicate the
        number of feasible trajectories for each method.
        Darker colors indicate lower optimization cost.
    }
    \label{fig:trajectory}

\end{minipage}
\vspace{-15pt}
\end{strip}
\par
\null
\clearpage